# Implementation And Performance Evaluation Of Background Subtraction Algorithms

Mr. Deepjoy Das and Dr. Sarat Saharia

Department of Computer Science & Engineering, Tezpur University, Assam, India

**ABSTRACT**

*The study evaluates three background subtraction techniques. The techniques ranges from very basic algorithm to state of the art published techniques categorized based on speed, memory requirements and accuracy. Such a review can effectively guide the designer to select the most suitable method for a given application in a principled way. The algorithms used in the study ranges from varying levels of accuracy and computational complexity. Few of them can also deal with real time challenges like rain, snow, hails, swaying branches, objects overlapping, varying light intensity or slow moving objects.*

## 1. INTRODUCTION

Background subtraction is basically detecting moving objects in videos using static cameras. The basic idea in the approach is detecting the moving objects from the difference between the current frame and a reference frame, which is called "background image" or "background model". The background image must be good enough to represent the scene with no moving objects and be regularly updated so that it adapt to the varying luminance conditions and geometry settings. Poor background image may result in poor background subtraction results, because it is to be subtracted with the current image to obtain the final result.

Processing a video stream to segment foreground objects from the background is a critical first step in many computer vision applications. The popularity of background subtraction algorithms largely comes from its computational efficiency, which allows applications such as human computer interaction, video surveillance and traffic monitoring to meet their real-time goals. Many different methods have been proposed over the recent years. Some of them are currently used in the CCTV detection application by the defense personals these different techniques varies in computational speed, memory requirements and accuracy basically.

There are many challenges faced during examining some of the techniques. Some of the challenges are shadows, wind causing cluttered areas. Traditional methods of background subtraction fails, if introduced with moving objects, reflections, illumination changes, Noise e.g. camera movement, object overlapping in the visual field, slow moving objects, objects being introduced or removed from the scene, movements through one or more above mentioned challenges and may produce inappropriate results. Robust, subtraction techniques are flexible enough to handle variations in lighting, moving scene clutter, multiple moving objects and other arbitrary changes to the observed scene. Such techniques are called real time background subtraction techniques.

## 2. BACKGROUND SUBTRACTION ALGORITHMS

We have implemented three background subtraction algorithms ranging from very basic strategy used to state of art techniques published. Some simple approaches aims to maximize speed and limits the memory requirements which produce a low accurate output such as the "frame difference" method [1] and other sophisticated approaches [2] [3] [5] aims to achieve the highest possible accuracy under possible circumstances.

### 2.1 FRAME DIFFERENCE TECHNIQUE THEORY [1]

Frame difference is the simplest form of background subtraction. The current frame is simply subtracted from the previous frame, and if the difference in pixel values for a given pixel is greater than a threshold $T_h$ then the pixel is considered part of the foreground [4].

$$|frame_i - frame_{i-1}| > T_h$$

The estimated background is just the previous frame and it is very sensitive to the threshold $T_h$.

Advantages of this method are it is a lowest computation speed of all the methods, background model is highly adaptive. Which means, this method adapt changes in the background faster than any other methods, since background is just the previous frame (precisely 1/fps). Disadvantages of this method are objects produced with uniformly distributed intensity values (such as the side of a car); the interior pixels can be interpreted as part of the background, objects must be continuously moving. If an object stays still for more than a frame period (1/fps), it becomes part of the background. This is an error.

The frame difference method also subtracts out background noise (such as waving trees), much better than the more complex approximate median and mixture of Gaussians methods A Challenge with this method is determining the threshold value. The result depends on threshold values thus each different video depends on different thresholds.

### 2.2 REAL TIME BACKGROUND SUBTRACTION AND SHADOW DETECTION TECHNIQUE THEORY [2]

The algorithms for real time background subtraction and shadow detection technique is published by T Horprasert, et al in the paper entitled "A statistical approach for real-time robust background subtraction and shadow detection" [2]. We briefly describe the method published in the paper by T Horprasert, et al [2]. This method is used for color image which has RGB components. A pixel $i$ in the image $\alpha_i = [\alpha_R(i), \alpha_G(i), \alpha_B(i)]$ [2] represent the pixel's expected RGB color value in the background and $\beta_i = [\beta_R(i), \beta_G(i), \beta_B(i)]$ [2] denote the pixel's RGB color value in a current image that is to be subtracted from the background. The main aims is to measure the distortion of $\beta_i$ from $\alpha_i$ which is done by breaking down the distortion into two parts, namely the brightness distortion [$\phi(\gamma_i) = (\beta_i - \gamma_i \alpha_i)^2$] [2] and chromaticity distortion [$\partial_i = \|\beta_i - \gamma_i \alpha_i\|$] [2]. Where $\gamma_i$ represents the pixel's strength of brightness with respect to the expected value. $\gamma_i = 1$ if the brightness of the given pixel in the current image is the same as in the reference image. $\gamma_i < 1$ if it is darker, and $\gamma_i > 1$ if it becomes brighter than the expected brightness. By applying suitable threshold on the brightness distortion $\phi(\gamma_i)$ and chromatic distortion $\partial_i$, we can distinguish the pixels as follows: original background (B) if it has both brightness and chromaticity similar to those of the same pixel in the background image. Shadow (S) if it has similar chromaticity but lower brightness than those of the same pixel in the background image.

## 2.3 ADAPTIVE BACKGROUND MIXTURE MODEL FOR REAL TIME TRACKING TECHNIQUE THEORY [3]

The algorithms for adaptive background mixture model for real time tracking technique is published by Stauffer, et al. in the paper entitled "Adaptive background mixture models for real-time tracking." [3]. we briefly describe the method mentioned in the paper by Stauffer, at al [3].

At first we need to know about pixel process. A pixel process of pixel $M$ is the values from frame 1 to frame $t$ also represented as $\{M_1,...M_j...M_t\}$ where $1 \leq j \leq t$. The probability of observing a pixel $M$ at frame $j$ denoted by $P(M_j) = \sum_{j=1}^{K} w_{j,t} \times \gamma(M_t, \mu_{j,t}, \Sigma_{j,t})$ [3], where, $\rho$ is the Gaussian probability density function $\gamma(M_t, \mu_{j,t}, \Sigma_{j,t}) = \frac{1}{\sqrt{|\Sigma|}\sqrt[n]{2\pi}} e^{-\frac{1}{2}(M_t-\mu_t)^T \Sigma^{-1}(M_t-\mu_t)}$, $\mu_{j,t}$ and $\Sigma_{j,t}$ are the mean and co-variance matrix of the $j^{th}$ Gaussian in the mixture at time $j$ respectively. $K$ is the no. of Gaussians in the mixture (proportional to memory available and computational power needed, usual value in 3-5). Covariance matrix is assumed to be of the form $\Sigma_{l,t} = \sigma_l^2 I$.

We update the probabilistic model every new pixel for each Gaussian in the mixture of pixel $M_t$. If $M_t \leq 2.5$ standard deviations from the mean then label 1 else labelled 0. For labeling of each Gaussian we have the following cases. I). If $\gamma_{j,t}$ is marked 1; then increase $w_{j,t}$, adjust $\mu_t$ closer to $M_t$ and decrease $\Sigma_t$. II). If the $\gamma_{j,t}$ is marked as 0; then decrease the $w_{j,t}$. III). If $\forall \gamma_{j,t}$ where $1 \leq j \leq t$ in the mixture for pixel $M_t$ are marked as 0 then Mark $M_t$ as a foreground pixel and find least probable Gaussian in the mixture and set $\mu_t = M_t$, $\sigma_t^2$ as high value, $w_{j,t}$ as a low value. For determining background pixel we look for distributions which have a high weight and low variance [3].

Update equations are as follows: $w_{j,t} = (1-\beta)*w_{j,t-1} + \beta*M_{j,t}$, $\mu_t = (1-\rho)*\mu_{t-1} + \rho*X_t$, $\sigma_t^2 = (1-\rho)*\sigma_{t-1}^2 + \rho*(X_t-\mu_t)^T*(X_t-\mu_t)$, $\rho = \beta*\gamma(X_t,\mu_{t-1},\Sigma_{t-1})$. $\beta$ is the learning parameter [3].

The methods has an advantage that it's robust against any background object if introduced with the walking subject such as rain, snow, handles motion from waving leaves and branches, noise

International Journal on Computational Sciences & Applications (IJCSA) Vol.4, No.2, April 2014

or working under day and night cycles. The disadvantages of this methods are when it's very windy and partly cloudy, if it's full sunny day because of long shadows (shadow can be classified as walking subject), varying illumination and if new object overlapping were some of the issues.

## 3. EXPERIMENTAL RESULTS

The data collection have been carried out with Nikon S3000 DLSR camera. QVGA (320×240) resolution videos has been captured and computations are performed using MATLAB version 7.8. For experiment video, five subjects were asked to walk at slow speed, normal speed and fast in front of the camera and these were shot in both indoor and outdoor settings. To test the efficiency of the algorithm in a proficient way, we have also introduced moving object and obstacles in few videos that forms the background of the walking subject. A total of 35 walking sequences were collected, between 5 to 7 sequences for each subject, averaging 15 to 35 seconds per subject. These videos were captured in 30 frames/seconds. Background subtraction of these video sequences is carried out using method "A statistical Approach for Real time Robust Background Subtraction and Shadow detection"[2] and "Adaptive background mixture models for real-time tracking"[3]. Video for method "frame difference" have been downloaded from internet since it requires appropriate lighting conditions and quality camera frames. Some of the results are shown below.

### 3.1 RESULT OF "FRAME DIFFERENCE" METHOD [1]

The result shows "the frame difference" [1] method as very low computationally intensive and efficient method. It also subtracts out background noise (such as waving trees), much better than the more complex approximate median and mixture of Gaussians methods (high computation methods). But the main challenge in this method is the determination of appropriate threshold, since the result solely depends on the threshold used.

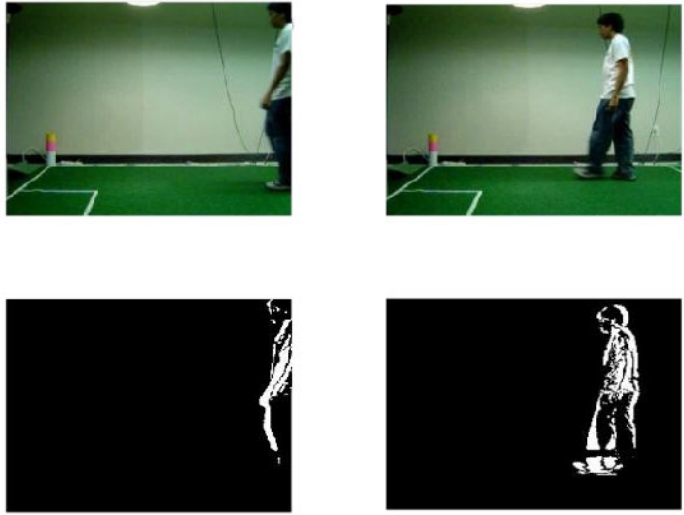

Fig 1: Result of frame difference background subtraction method. Low accuracy and low computational intensive.

## 3.2 RESULT OF "A STATISTICAL APPROACH FOR REAL TIME ROBUST BACKGROUND SUBTRACTION AND SHADOW DETECTION" METHOD[2]

This method produces average result with average computation efficiency among the other two methods, it classifies background and shadow according to two computation models i.e. Brightness distortion and colour distortion. So, the methods solely depend on the each pixels colour value, each pixels colour value should be precise enough to represent itself in one of the component of the result. So, the image has to be of good quality, possibly greater than QVGA resolution for better output. The experiment was performed on QVGA resolution, the result (some of which are given below) is not that efficient, but it compensates on computational efficiency.

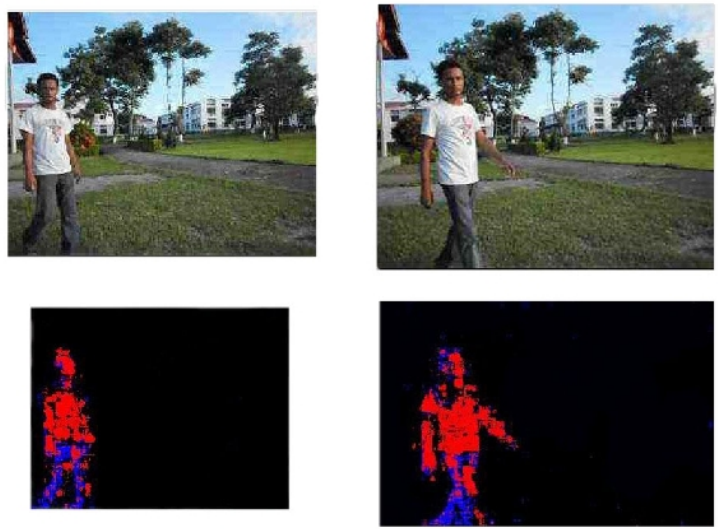

Fig 2: Result of background subtraction using statistical approach for real time robust background subtraction and shadow detection technique. Red coloured part denotes foreground object and blue coloured part denotes shadow. Medium accuracy and medium computational intensive.

## 3.3 RESULT OF "ADAPTIVE BACKGROUND MIXTURE MODELS FOR REAL-TIME TRACKING" METHOD [3]

This method has the best result among all the above method discussed because it can deal with slow lighting changes and other challenges. It also deals with multi-modal distributions caused by moving branches of tress, snow falls, shadows, flying birds and other troublesome features of the real world, but it also costs the maximum in computational time and memory. The computation time mainly depend on number of Gaussian components (Gaussian distribution) considered for Mixture of Gaussian (MOG) models, which typically ranges from 3 to 5. In the experiment number of component taken is 3, one can take it as 5 depending on multimodal density required. Some of the experimental results are shown below.

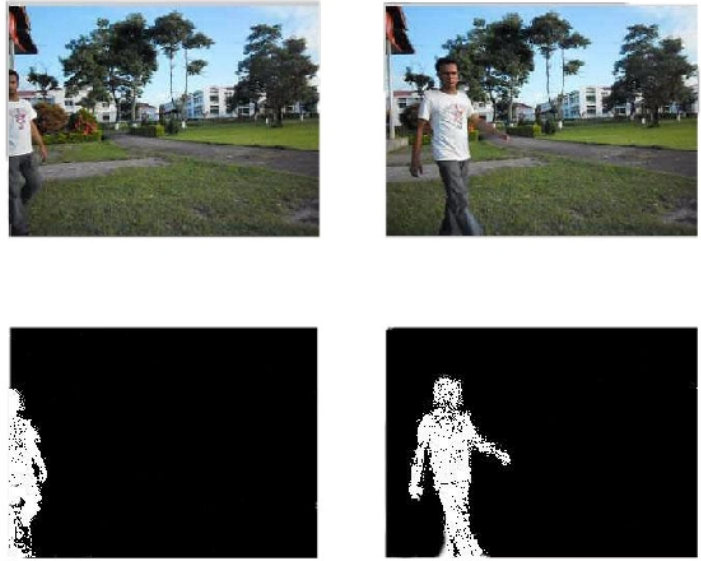

Fig 3: Result of background subtraction using adaptive background mixture models for real time tracking. High accuracy and high computational intensive.

## 4. CONCLUSION

The results for some of the video are imperfect due to camera noise and appropriate lighting condition. The result of "Adaptive background mixture models for real-time tracking " and "A statistical Approach for Real time Robust Background Subtraction and Shadow detection" does not depends on lighting condition and some of the mentioned challenges though it depends on quality of the image to an extent. The result of frame difference depends on lighting conditions etc. which is the reason a different video is taken to produce the result (the video was downloaded from internet), as the frame difference methods subtracts background frame from the current frame, the background is just the previous frame, It produce an erroneous result due to slow moving objects, lighting condition and many other challenges if introduced with.

The result of the A Statistical Approach for Real-time Robust Background Subtraction and Shadow Detection" [2] and "Adaptive background mixture models for real-time tracking"[3] papers can be used for real time background subtraction method, but the result of mixture of Gaussian method [3]  is computationally intensive. The most appropriate method is [2] due to average computation cost and average result.

**AUTHORS**

**Dr. Sarat Saharia**
He is currently working as Associate Professor in the department of computer science and engineering at Tezpur University, Assam, India. His research interest includes pattern recognition and image processing

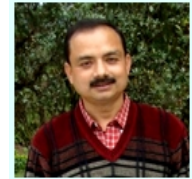

**Deepjoy Das**
Has done Bachelor of Technology from Tezpur University and was held as Assistant Project Engineer in Indian Institute of Technology for one year. His research interest includes image processing with supervised and unsupervised learning and handwriting recognition.

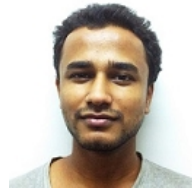